\documentclass{article}

\usepackage{w2rep_preprint,times}

\usepackage[utf8]{inputenc}
\usepackage[T1]{fontenc}
\usepackage{booktabs}
\usepackage{amsfonts}
\usepackage{amssymb}
\usepackage{amsmath}
\usepackage{nicefrac}
\usepackage{microtype}
\usepackage{graphicx}
\usepackage{adjustbox}
\usepackage{doi}
\usepackage{xcolor}
\usepackage{xspace}
\usepackage{url}
\usepackage{hyperref}
\usepackage{cleveref}

\definecolor{w2blue}{HTML}{3F6FB5}
\definecolor{w2teal}{HTML}{238B8E}
\definecolor{w2red}{HTML}{C94C4C}

\newcommand{\ours}{\textsc{W2Rep}\xspace}

\title{W2Rep: Learning Visual Representations by Watching the World Change}

\author{
Wen Huang\textsuperscript{\rm 1,*} \quad
Hang Guo\textsuperscript{\rm 1,*} \quad 
Jiarui Yang\textsuperscript{\rm 2} \quad
Zheng Liu\textsuperscript{\rm 1} \quad
Tao Dai\textsuperscript{\rm 3,$\dagger$} \quad
Shu-Tao Xia\textsuperscript{\rm 1} \\
\textsuperscript{\rm 1}Tsinghua Shenzhen International Graduate School,
Tsinghua University, Shenzhen, China \\
\textsuperscript{\rm 2}Nankai University, Tianjin, China \quad
\textsuperscript{\rm 3}Shenzhen University, Shenzhen, China \\
\textsuperscript{*}Equal contribution. \quad
\textsuperscript{$\dagger$}Corresponding author. \\
\texttt{huang-w24@mails.tsinghua.edu.cn} \quad
\texttt{daitao.edu@gmail.com}
}

\iclrfinalcopy

\hypersetup{
pdftitle={W2Rep: Learning Visual Representations by Watching the World Change},
pdfauthor={Wen Huang, Hang Guo, Jiarui Yang, Zheng Liu, Tao Dai, Shu-Tao Xia},
pdfkeywords={self-supervised learning, visual representation, clip representation, temporal supervision, masked prediction, clip conditioning, JEPA},
}

\begin{document}
\maketitle

\begin{abstract}
Images capture the world at one moment, whereas video reveals how it changes.
Image self-supervision learns spatial structure from a single moment, while
video methods commonly learn temporal relationships inside a representation
computed jointly from several frames. We ask whether watching a scene change
can instead improve features available from one image without sacrificing the
ability to represent video. We introduce \ours{}, a masked
feature-prediction framework in which an independently encoded source image
participates in prediction at the same or another moment. The predictor is
conditioned on visible video context, the queried location, and the signed time
interval between source and target. This gives the cross-frame objective two
complementary roles: the image path learns features that remain useful across
time, while the video path must gather evidence that is missing from the
source image. Across model scales and downstream tasks, \ours{} improves frozen and
fine-tuned recognition under our comparison protocol, while joint video
encoding provides further gains over frame-wise aggregation. Controlled
experiments show that these gains depend on directly updating the source-image
features and on using both video context and temporal displacement. Overall,
change across a video can supervise a visual encoder whose representations
remain useful at either image or video granularity. Code is available at~\href{https://wenooi.github.io/W2Rep}{https://wenooi.github.io/W2Rep}.
\end{abstract}

\section{Introduction}

Learning visual representations that transfer across tasks and input formats
is a central goal of visual pretraining. A useful representation should capture
not only what is visible in one image, but also how related observations of the
same scene are organized as the world changes. Images provide spatial
structure at a single moment; video additionally connects objects, states, and
interactions across time. These temporal relations offer supervision that is
unavailable from isolated images.

Self-supervised learning has progressively expanded the relationships used to
train visual representations. Contrastive learning
\citep{chen2020simple,he2020momentum,caron2020unsupervised} and self-distillation
\citep{grill2020bootstrap,caron2021emerging} relate augmented views of one
image, while masked modeling predicts hidden pixels, tokens, or features from
visible image context \citep{bao2021beit,he2022masked,assran2023self}. Video
methods extend these ideas across space and time, commonly encoding several
frames together to learn a spatiotemporal representation
\citep{wei2022masked,tong2022videomae,feichtenhofer2022masked,bardes2024revisiting}.
These approaches have produced strong image and video models, but they differ
in whether the encoder is trained to represent one image or an entire clip.

\begin{figure*}[t]
    \centering
    \includegraphics[width=0.94\textwidth]{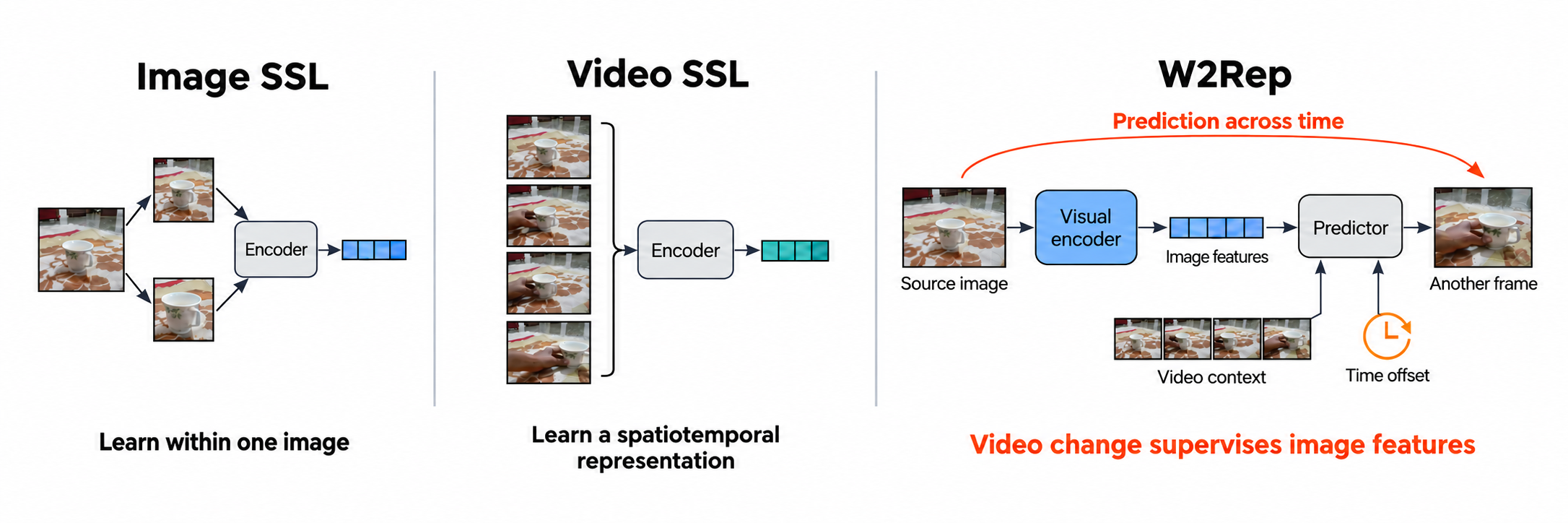}
    \caption{Three training interfaces for visual self-supervision. Image
    methods learn from different views or masked regions of one image. Video
    methods commonly encode several frames together. \ours{} uses change across
    a video to train image features. Its video-context path supplies the
    complementary multi-frame evidence needed for cross-frame prediction, so
    the same objective also trains the encoder to use video inputs. The retained
    encoder can later process either an image or a video. The thumbnails show four frames from a
    single Something-Something V2 video; the illustration is schematic rather
    than an exhaustive taxonomy.}
    \label{fig:training-paradigms}
\end{figure*}

Many existing video objectives are designed primarily to learn representations
of clips rather than individual images. Because they process several frames
together, the resulting representation can use information from the entire
clip. This is valuable for video understanding, but it does not directly answer
whether observing a scene change can also improve the representation produced
from one image alone. As illustrated in Fig.~\ref{fig:training-paradigms}, we
study this complementary question: can changes across frames help learn better
image representations, while still allowing the model to use multiple frames
when video is available?

Our idea is to make an independently encoded source image participate in
prediction across time. Because the scene may change from one frame to another,
we do not directly match their features, as is commonly done between two
augmented views of the same image. Instead, the source image helps predict the
features of a queried region at the same or another time. The prediction also
uses the queried location, the time interval, and visible evidence from the
video. This relates observations in one learned feature space without imposing
temporal invariance or defining change through a hand-crafted target such as
optical flow or an RGB difference.

\ours{} realizes this idea with a visual encoder and a conditional predictor
used only during pretraining, as detailed in Fig.~\ref{fig:w2rep-overview}.
The same visual encoder processes the sampled source image and the masked video
clip in two separate passes. The first produces image features without seeing
neighboring frames; the second summarizes visible context from the clip. Given
the source features, clip context, masked spatial queries, and a signed temporal offset,
the predictor estimates target features in the source or another frame.
Both passes are necessary for cross-frame prediction: the source-image pass
must produce features from one image that remain useful for predicting other
moments, while the video pass gathers complementary evidence from the
surrounding clip. The cross-frame loss therefore trains the same Vision
Transformer (ViT) both as an
image encoder and as a multi-frame encoder. After pretraining, the predictor
and auxiliary tokens are discarded.
The retained encoder produces image-level features from one image and can also
aggregate evidence jointly when given several frames. We call these features a
\emph{temporally grounded visual state}: they remain available from an image,
while the distinctions they preserve are learned from related observations of
scenes as they change.

\ours{} improves frozen recognition at both ViT-B/16 and ViT-L/16 scales while
also producing effective spatiotemporal representations. At ViT-B/16, frozen
ImageNet accuracy rises from $28.5\%$ for compute-matched I-JEPA to $34.6\%$. Across
Something-Something V2 (SSv2), UCF101, and Diving48, encoding frames together
consistently improves over
aggregating independently encoded frames, and full fine-tuning reaches
$58.8\%$ on SSv2, compared with $55.2\%$ for step-matched VideoMAE. Ablations
show complementary roles for same-frame and cross-frame prediction and verify
that both visible clip context and signed temporal displacement affect the
prediction. A frozen-predictor diagnostic further retrieves the requested
target frame in $52.1\%$ of eight-way comparisons, versus $12.5\%$ at random.

Our contributions are:
\begin{itemize}
    \item We introduce a masked feature-prediction objective that places an
    independently computed image representation in both within-frame and
    cross-frame prediction.

    \item We realize this objective with one ViT backbone that produces both
    image and spatiotemporal video representations after its auxiliary
    prediction components are removed.

    \item Across two model scales, frozen and fine-tuned evaluations, and
    controlled temporal diagnostics, we show that this supervision improves
    image- and video-level recognition and that the predictor uses both ordered
    clip evidence and signed temporal displacement.
\end{itemize}

\section{Related Work}
\paragraph{Image representation learning.}
Image self-supervision learns visual structure through view matching,
self-distillation, and masked prediction. SimCLR and MoCo
\citep{chen2020simple,he2020momentum} contrast representations of augmented
views; SwAV \citep{caron2020unsupervised} performs online clustering; and BYOL and DINO
\citep{grill2020bootstrap,caron2021emerging} learn from slowly updated or
self-distilled targets. VICReg \citep{bardes2021vicreg} instead controls
collapse through a feature-statistics objective. BEiT and MAE
\citep{bao2021beit,he2022masked} predict discrete tokens or pixels, while iBOT
and data2vec \citep{zhou2021ibot,baevski2022data2vec} combine masking with
learned target representations. I-JEPA \citep{assran2023self} likewise moves
prediction to feature space, matching target-region representations from
visible image context. \ours{} builds on this
latent-prediction principle. Its zero-offset task provides a within-frame
completion constraint, while its displaced targets use relations between
observations that are unavailable to an image-only objective. This distinction
does not imply that image methods cannot learn action-relevant features; it
concerns the source of their pretraining supervision.

\paragraph{Masked and predictive video learning.}
Masked video objectives reconstruct pixels
\citep{tong2022videomae,feichtenhofer2022masked,girdhar2023omnimae}, predict
discrete visual tokens \citep{wang2022bevt}, hand-crafted features
\citep{wei2022masked}, or teacher features
\citep{wang2023masked,bardes2024revisiting}. VideoMAE~V2
\citep{wang2023videomae} adds decoder-side masking to scale pixel
reconstruction, while V-JEPA \citep{bardes2024revisiting} uses a
spatiotemporal encoder throughout pretraining and downstream clip inference.
V-JEPA~2.1
\citep{mur2026v} extends this formulation by supervising visible
predictor tokens and multiple encoder depths, which its experiments show is
important for dense features. These methods demonstrate the strength of joint
clip representations. \ours{} addresses a complementary interface: an
explicit single-frame source path participates in cross-frame prediction, and
the same retained patch encoder can later be called on either one frame or a
jointly encoded clip. Its final objective supervises only sampled mask queries
at the last target layer, so we do not claim the dense supervision provided by
V-JEPA~2.1.

\paragraph{Frame representations from temporal prediction.}
Several works more directly use change across frames to train image- or
frame-compatible representations. Video contrastive objectives align augmented
clips \citep{qian2021spatiotemporal}, match short and long temporal views
\citep{wang2022long}, or learn from playback speed
\citep{benaim2020speednet}. Other objectives verify temporal order
\citep{misra2016shuffle} or learn correspondence through temporal cycle
consistency and contrastive walks
\citep{wang2019learning,dwibedi2019temporal,jabri2020space}.
\citet{pathak2017learning} use motion-based
segmentation as pseudo-label supervision for a network that
segments objects from a single frame. RSP \citep{jang2024rsp} addresses the
ambiguity of future-frame prediction with a stochastic pixel-generation model
and adds masked image modeling for within-frame information. MC-JEPA
\citep{bardes2023mc} jointly learns content features and dense optical flow
with one backbone. TDV
\citep{daithankar2026you} predicts an additive next-frame latent update from an
RGB temporal difference processed by a separate motion encoder. \ours{} predicts
masked exponential-moving-average (EMA) features at one
zero offset and multiple positive or negative offsets, conditioning on visible
context from a masked clip. It requires neither motion-segmentation
pseudo-labels, stochastic pixel generation, an optical-flow target, an explicit
RGB difference, nor a dedicated temporal-difference encoder.

\paragraph{Compressed temporal variables and latent actions.}
ToBo \citep{kim2026token} compresses a reference scene into a single bottleneck
token and uses that token with a small number of target-scene patches to
reconstruct the subsequent scene. Its bottleneck is trained as a compact scene
representation. By contrast, the auxiliary clip latents in \ours{} are inferred
from the visible context of the masked clip, condition the predictor together
with a separately encoded source frame, and are discarded after pretraining.
Latent-action world models pursue a different endpoint: for example, Genie
\citep{bruce2024genie} learns latent actions for controllable generation. Since
\ours{} provides neither an action mapping nor a control experiment, we treat
its auxiliary latents only as a predictive condition and do not identify them
as a bottleneck, a pure temporal variable, or a latent action.

\section{Method}
\label{sec:method}
\begin{figure*}[t]
    \centering
    \includegraphics[width=0.96\textwidth]{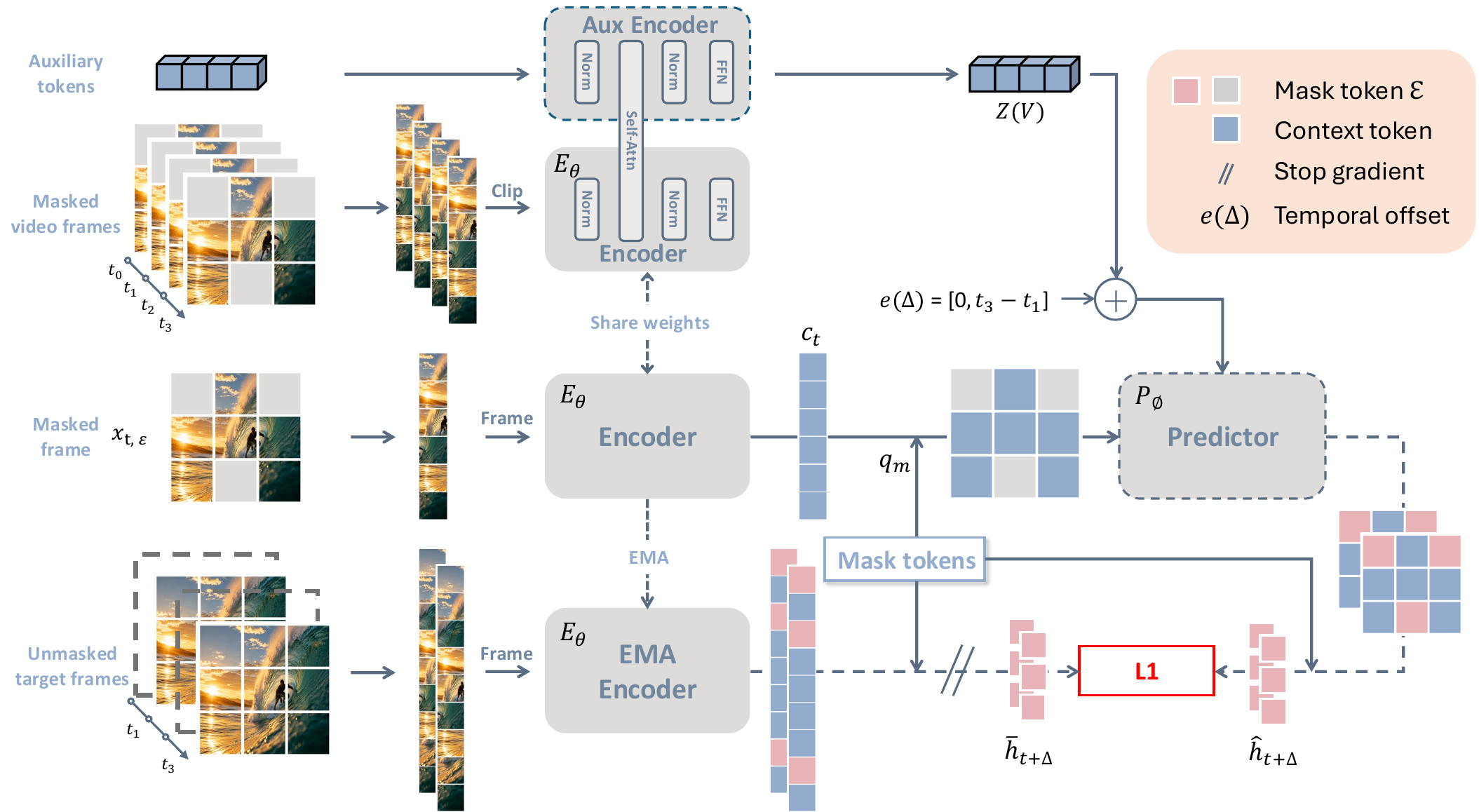}
    \caption{Overview of \ours{}. The same student visual encoder separately
    processes a masked source image and a masked video. A predictor combines
    the resulting source-image features with visible video context, masked
    spatial queries, and a signed temporal offset to predict features produced
    by a slowly updated target encoder. After pretraining, only the visual
    encoder is retained for image or video inference.}
    \label{fig:w2rep-overview}
\end{figure*}

As shown in Fig.~\ref{fig:w2rep-overview}, \ours{} trains image features through masked prediction within an image and
across time. 
We first
describe the two encoding passes and exponential-moving-average (EMA) targets
(Section~\ref{sec:method-encoding}), then the spatial masks and target sampling
(Section~\ref{sec:method-sampling}), and finally the predictor and training
objective (Section~\ref{sec:method-prediction}).

\subsection{Encoding the source image and video context}
\label{sec:shared-encoder}
\label{sec:method-encoding}
The two encoding passes separate the image representation we want to keep from
the video context used only during pretraining. Let $\mathcal E$ denote the
visible spatial locations defined in Section~\ref{sec:method-sampling}, and let
$d$ denote the encoder feature dimension. The student ViT $E_\theta$ is applied
twice to each training example. The source-image pass produces visible image
features $c_t$. The video pass produces patch features $H^V$ and the final
states $z(V)$ of $K$ auxiliary tokens, each with feature dimension $d$:
\begin{align}
    c_t &= E_\theta(x_{t,\mathcal E};\varnothing),
    \label{eq:source-context}\\
    (H^V,z(V)) &= E_\theta(V_{\mathcal E};z_0),
    & z(V)&\in\mathbb{R}^{K\times d}. \label{eq:clip-latents}
\end{align}
The source pass produces the image features used for prediction. Specifically,
$c_t$ contains the visible source-image features, and $\varnothing$ means that
no auxiliary tokens are used in this pass. The video pass begins with $K$
learned tokens $z_0$ and returns their final states $z(V)$. We use $K{=}16$.
Both passes use the same fixed, separable spatiotemporal sinusoidal position
encoding. An image is treated as a one-frame sequence at temporal position
zero. The auxiliary tokens $z_0$
receive no position encoding. Standard bidirectional attention allows these
tokens to gather context from the visible video patches. Only $z(V)$ is given
to the predictor; the video patch output $H^V$ is not otherwise used.

The video-context path provides the information that is missing from the
source image. Cross-frame targets generally cannot be predicted from $c_t$
alone, and $z(V)$ is the predictor's only input that contains evidence from
the other visible frames. Reducing the cross-frame loss therefore requires
$E_\theta$ to aggregate useful multi-frame evidence into $z(V)$. The same
$z(V)$ is reused for every target in a training sample, and the video pass is
not told which frame is the source, which frames will be targets, or which
offsets will be queried. It must consequently summarize context useful across
several possible predictions rather than encode a target-specific answer.

This requirement also explains why pretraining benefits video inference even
though the prediction targets are frame features. Gradients from every
cross-frame prediction pass through $z(V)$ into the masked multi-frame forward
of $E_\theta$, training its attention layers to collect evidence across the
clip. Those same layers process the video patch tokens at downstream time.
The auxiliary tokens are no longer needed after pretraining.

Following common joint-embedding practice
\citep{grill2020bootstrap,caron2021emerging,assran2023self}, the EMA copy of the
student encoder provides stable prediction targets. We denote its parameters
by $\bar\theta$. For a frame at temporal offset $\Delta$ from the source, it
encodes the complete target frame without auxiliary tokens and applies layer
normalization (LN):
\begin{equation}
    \bar h_{t+\Delta}
    =\operatorname{LN}\!\left(E_{\bar\theta}
      (x_{t+\Delta};\varnothing)\right),
    \label{eq:target}
\end{equation}
Gradients are stopped at $\bar h_{t+\Delta}$, and $\bar\theta$ is updated as an
exponential moving average of $\theta$.

\subsection{Masking and target sampling}
\label{sec:method-sampling}
A common spatial mask gives every target time the same set of prediction
locations. Following I-JEPA-style block sampling \citep{assran2023self}, we
sample target blocks with concatenated position list $\mathbf m$ and select the
visible encoder locations $\mathcal E$ from outside those blocks. Following
the tube-masking strategy commonly used in video self-supervised learning
\citep{tong2022videomae,bardes2024revisiting}, the same $\mathcal E$ and
$\mathbf m$ are used in every frame. These shared locations provide consistent
spatial queries across time; they do not assume that an object remains at the
same location. Target blocks may overlap, so $\mathbf m$ is a list of queries
rather than a partition of the image.

Temporal sampling determines which moments are predicted from each source
image. We sample the source index $t$ uniformly and choose $M$ distinct target
frames without replacement from the remainder of the video. Their signed time
offsets $\{\Delta_i\}_{i=1}^{M}$ include frames before and after the source. We
also include $\Delta{=}0$ for masked completion within the source image, giving
$\mathcal D=\{0\}\cup\{\Delta_i\}_{i=1}^{M}$. Exact masking and sampling
hyperparameters are given in Appendix~\ref{app:implementation-details}.

\subsection{Prediction and training objective}
\label{sec:method-prediction}
Each prediction query specifies both where and when to predict. The spatial
queries $q_{\mathbf m}$ use one learned mask-token initialization and
two-dimensional position encodings to distinguish locations in $\mathbf m$.
We denote the predictor by $P_\phi$. For each target time
$\Delta\in\mathcal D$, it computes
\begin{equation}
    z_\Delta=\mathbf{1}[\Delta\neq0]z(V), \qquad
    \hat h_{t+\Delta}^{\mathbf m}
      =P_\phi\!\left(c_t,q_{\mathbf m},z_\Delta,e(\Delta)\right),
    \label{eq:pred}
\end{equation}
Here, $\mathbf 1[\cdot]$ is the indicator function, and $e(\Delta)$ is a
learned transformation of a sinusoidal encoding
\citep{vaswani2017attention} of the
signed time offset, whose sign distinguishes frames before and after the
source. For $\Delta{=}0$, we set $z_\Delta$ to zero so that same-frame
completion cannot use video context. The same predictor therefore handles both
same-frame and cross-frame prediction.

The predictor keeps location, time, and video evidence as distinct inputs. It
projects $c_t$, adds the source positions, and concatenates the result with
$q_{\mathbf m}$. Each predictor block uses an offset-conditioned
cross-attention branch to read $z_\Delta$, followed by self-attention and a
multilayer perceptron (MLP) over the source and query tokens. Only the
query-token outputs are
projected back to the encoder dimension. Thus, $q_{\mathbf m}$ specifies where
to predict, $e(\Delta)$ specifies when to predict, and $z(V)$ supplies visible
evidence from the video.

Training aligns each prediction with the normalized EMA feature at the
requested location and time:
\begin{equation}
    \mathcal L_\Delta
      =\operatorname{SmoothL1}\!\left(
        \hat h_{t+\Delta}^{\mathbf m},
        \operatorname{gather}(\bar h_{t+\Delta},\mathbf m)
      \right).
    \label{eq:loss}
\end{equation}
The final objective gives equal weight to same-frame completion and each of the
$M$ cross-frame predictions, and adds a weak scale regularizer on $z(V)$:
\begin{equation}
    \mathcal L
      =\frac{1}{M+1}\sum_{\Delta\in\mathcal D}\mathcal L_\Delta
       +\lambda_z\operatorname{mean}\!\left(z(V)^2\right),
    \qquad \lambda_z=10^{-4}.
    \label{eq:total-loss}
\end{equation}

\section{Experiments}
\label{sec:experiments}
Our experiments center on one question: does cross-frame prediction produce a
visual encoder that is useful when given either an image or a video?
Section~\ref{sec:experimental-setup} defines the comparison budgets, downstream
tasks, and the two input readouts. Section~\ref{sec:frozen-transfer} evaluates
the resulting representations across tasks and model scales, including
end-to-end adaptation. Section~\ref{sec:controlled-ablation} then isolates the
training signals responsible for the observed behavior and examines how the
final model uses temporal conditioning. Additional protocols and analyses are
provided in the appendix.

\subsection{Experimental setup}
\label{sec:experimental-setup}
\paragraph{Pretraining data and comparison budgets.}
We pretrain all models from scratch on SSv2 at $224{\times}224$ resolution and
evaluate only the retained encoder. Two pre-specified controls reflect the
methods' different training units. Video-native methods share a 100k-update
horizon, global batch 256, and eight-frame, stride-three clips, matching nominal
clip/frame exposure and update count. Image- or frame-native methods consume
different numbers of frames and model calls per update, so we instead match
their total profiled forward multiply--accumulate operations (MACs) to \ours{}
at each backbone scale. The profiles include all training-time branches, while
baselines retain their method-specific objectives and optimization. These
family-level controls do not assert identical wall-clock or complete training
cost; component-level conclusions come only from the controlled ablations in
Section~\ref{sec:controlled-ablation}. Appendix~\ref{app:implementation-details}
provides the accounting and an alternative compute-matched VideoMAE comparison.

\paragraph{Tasks and metrics.}
We evaluate frozen representations on ImageNet-1K classification
\citep{deng2009imagenet,russakovsky2015imagenet}, ADE20K semantic segmentation
\citep{zhou2017scene}, and action recognition on SSv2
\citep{goyal2017something}, UCF101 \citep{soomro2012ucf101}, and Diving48
\citep{li2018resound}.
Classification numbers with $\pm$ are means over three fixed-schedule probe
seeds; ADE20K trains a UPerNet decoder \citep{xiao2018unified} on a frozen
backbone. We
additionally fine-tune the full
encoder on SSv2 under one shared downstream recipe. ViT-B/16 is our primary
setting and ViT-L/16 tests scaling.

\paragraph{Image and video readouts.}
For action recognition, \emph{independent-$8$} averages features from eight
separate frame calls, while \emph{joint-$8$} gives the same frames to one
spatiotemporal encoder call. The first readout uses the same single-image path
that participates in every pretraining example. The second lets the retained
ViT process multiple frames together; this patch-only multi-frame input is
supported by its attention and position encodings but is not an additional
pretraining branch. Detailed schedules, budget accounting, baseline
adaptations, and evaluation protocols appear in
Appendix~\ref{app:implementation-details}.

\subsection{Frozen transfer across tasks and scales}
\label{sec:frozen-transfer}
Frozen transfer provides the primary test of whether pretraining benefits both
image and video readouts. Table~\ref{tab:main-transfer} evaluates the same
encoders first on image-oriented tasks and then through frame-wise and
spatiotemporal action readouts. Each row uses one primary pretraining checkpoint.

\begin{table*}[t]
\centering
\scriptsize
\setlength{\tabcolsep}{4.0pt}
\renewcommand{\arraystretch}{1.10}
\caption{Frozen transfer at two backbone scales. ImageNet-1K and action
columns report top-1 accuracy (\%); ADE20K reports mean intersection-over-union
(mIoU). \emph{Ind.} averages
eight independently encoded frames, whereas \emph{joint} applies space--time
attention to the same eight frames. Values with $\pm$ average three probe
seeds. Bold denotes the best result within each backbone and readout. A dash
indicates that the readout is not applicable or was not evaluated.}
\label{tab:main-transfer}

\textbf{(a) Image-oriented transfer}\par\smallskip
\begin{tabular}{lcc}
\toprule
Method & ImageNet-1K & ADE20K \\
\midrule
\multicolumn{3}{l}{\textbf{ViT-B/16}} \\
I-JEPA \citep{assran2023self}       & $28.50{\pm}0.12$                & $20.55$ \\
VideoMAE \citep{tong2022videomae}  & $26.70{\pm}0.10$       & $\mathbf{25.79}$ \\
V-JEPA \citep{bardes2024revisiting}& $24.90{\pm}0.09$       & $21.97$ \\
TDV \citep{daithankar2026you}
                                    & $7.29{\pm}0.15$        & $14.05$ \\
RSP \citep{jang2024rsp}             & $24.76{\pm}0.07$       & $17.84$ \\
\ours{}                             & $\mathbf{34.60{\pm}0.05}$ & $22.21$ \\
\addlinespace[3pt]
\multicolumn{3}{l}{\textbf{ViT-L/16}} \\
I-JEPA \citep{assran2023self}
                                    & $31.27{\pm}0.07$       & $21.67$ \\
VideoMAE \citep{tong2022videomae}  & $30.68{\pm}0.11$       & $\mathbf{27.13}$ \\
V-JEPA \citep{bardes2024revisiting}& $24.16{\pm}0.17$       & $23.91$ \\
\ours{}                             & $\mathbf{35.37{\pm}0.12}$ & $23.62$ \\
\bottomrule
\end{tabular}

\vspace{0.7em}
\textbf{(b) Action recognition}\par\smallskip
\resizebox{\textwidth}{!}{%
\begin{tabular}{lcccccc}
\toprule
& \multicolumn{2}{c}{\textbf{SSv2}} &
  \multicolumn{2}{c}{\textbf{UCF101}} &
  \multicolumn{2}{c}{\textbf{Diving48}} \\
\cmidrule(lr){2-3}\cmidrule(lr){4-5}\cmidrule(lr){6-7}
Method & Ind.-8 & Joint-8 & Ind.-8 & Joint-8 & Ind.-8 & Joint-8 \\
\midrule
\multicolumn{7}{l}{\textbf{ViT-B/16}} \\
I-JEPA \citep{assran2023self}
& $8.24{\pm}0.04$ & -- & $47.69{\pm}0.30$ & -- & $9.02{\pm}0.11$ & -- \\
VideoMAE \citep{tong2022videomae}
& $8.98{\pm}0.04$ & $24.27{\pm}0.07$ & $46.29{\pm}0.15$ & $53.87{\pm}0.19$ & $8.00{\pm}0.11$ & $10.24{\pm}0.33$ \\
V-JEPA \citep{bardes2024revisiting}
& $6.77{\pm}0.16$ & $12.66{\pm}0.04$ & $42.78{\pm}0.12$ & $49.23{\pm}0.44$ & $8.38{\pm}0.09$ & $8.80{\pm}0.36$ \\
TDV \citep{daithankar2026you}
& $2.21{\pm}0.05$ & -- & $20.13{\pm}0.37$ & -- & $6.70{\pm}0.38$ & -- \\
RSP \citep{jang2024rsp}
& $10.20{\pm}0.05$ & -- & $44.85{\pm}0.22$ & -- & $8.43{\pm}0.38$ & -- \\
\ours{}
& $\mathbf{12.64{\pm}0.03}$ & $\mathbf{25.38{\pm}0.10}$ & $\mathbf{51.40{\pm}0.28}$ & $\mathbf{56.60{\pm}0.38}$ & $\mathbf{10.08{\pm}0.26}$ & $\mathbf{11.29{\pm}0.25}$ \\
\addlinespace[3pt]
\multicolumn{7}{l}{\textbf{ViT-L/16}} \\
I-JEPA \citep{assran2023self}
& $9.44{\pm}0.06$ & -- & $49.75{\pm}0.25$ & -- & $10.73{\pm}0.08$ & -- \\
VideoMAE \citep{tong2022videomae}
& $11.57{\pm}0.12$ & $29.57{\pm}0.20$ & $49.95{\pm}0.28$ & $58.49{\pm}0.07$ & $8.98{\pm}0.22$ & $9.83{\pm}0.46$ \\
V-JEPA \citep{bardes2024revisiting}
& $6.68{\pm}0.12$ & $11.87{\pm}0.12$ & $41.51{\pm}0.19$ & $47.49{\pm}0.28$ & $9.07{\pm}0.21$ & $10.54{\pm}0.58$ \\
\ours{}
& $\mathbf{14.30{\pm}0.17}$ & $\mathbf{31.91{\pm}0.08}$ & $\mathbf{52.74{\pm}0.04}$ & $\mathbf{58.94{\pm}0.07}$ & $\mathbf{10.88{\pm}0.16}$ & $\mathbf{11.81{\pm}0.51}$ \\
\bottomrule
\end{tabular}%
}
\end{table*}

\paragraph{W2Rep improves image-level transfer.}
Under the stated comparison protocol, \ours{} gives the strongest ImageNet
and independent-frame action transfer at both backbone scales. In particular,
the independent SSv2 result improves from $8.24\%$ for compute-matched I-JEPA
to $12.64\%$ for \ours{} at ViT-B. 

\begin{figure*}[t]
    \centering
    \includegraphics[width=0.96\textwidth]{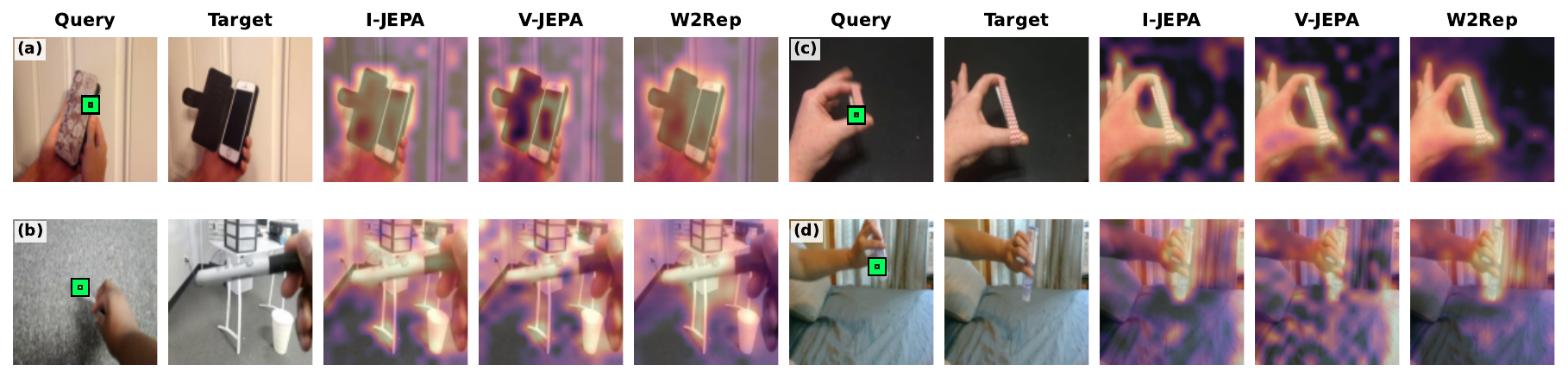}
    \caption{Qualitative cross-frame patch similarity for four SSv2 examples,
    arranged as (a--b) on the left and (c--d) on the right. Query and target
    frames are encoded independently, without access to temporal context. The
    green box marks a $16\!\times\!16$ query patch in the source frame. Each
    heatmap shows the mean-centered cosine similarity between that query and
    the final-layer target-frame patch features. Colors are normalized within
    each map using its 5th and 95th similarity percentiles and therefore
    indicate spatial structure, not similarity magnitudes across models.}
    \label{fig:cross-frame-patch-similarity}
\end{figure*}

\paragraph{Cross-frame patch similarity from frame-only features.}
Figure~\ref{fig:cross-frame-patch-similarity} provides a local view of the
retained representation. We select a patch on the manipulated object in one
frame and compare its feature with every patch in a later frame, while encoding
the two frames independently. Across changes in pose and configuration,
\ours{} retains spatially coherent similarity over the related object or
interaction region. In these examples, its response is also more concentrated
on the relevant region and less diffuse over the background than the I-JEPA
and V-JEPA maps. The baselines nevertheless preserve useful correspondence in
some cases. We treat this pattern as qualitative: the visualization
complements the recognition results by showing the local structure of the
frame features, rather than establishing a quantitative tracking advantage.

\paragraph{Joint encoding adds complementary temporal information.}
When the same eight frames are encoded together, \ours{} improves from
$12.64\%$ to $25.38\%$ on SSv2 and from $51.40\%$ to $56.60\%$ on UCF101.
The same pattern holds after scaling to ViT-L. This video capability is also
trained by the cross-frame objective: predicting another frame requires
$z(V)$ to carry evidence gathered from the masked multi-frame input, so the
loss backpropagates through the encoder's video pass rather than only through
the independently encoded source image. The joint-readout gains are consistent
with this mechanism: after the training-time context tokens are removed, the
same attention layers can still combine evidence across the input frames.

\paragraph{Scaling transfers, but dense localization remains a boundary.}
Moving from ViT-B to ViT-L improves every reported \ours{} readout. The pattern
is less favorable on ADE20K: \ours{} rises from $22.21$ to $23.62$ mIoU, but
VideoMAE remains stronger at both scales ($25.79$ and $27.13$). Our evidence
therefore supports recognition transfer and scaling, not a general advantage
for dense prediction.

\paragraph{End-to-end fine-tuning.}
The advantage is retained when the full encoder is adapted to SSv2: \ours{}
reaches $58.77\%$ top-1, compared with $55.16\%$ for VideoMAE under the same
50-epoch fine-tuning recipe. Table~\ref{tab:ssv2-fullft} and the complete
protocol are provided in Appendix~\ref{app:implementation-details}.

\subsection{Ablation Study}
\label{sec:controlled-ablation}
Table~\ref{tab:controlled-ablation} tests the prediction objectives, temporal
offset, video context, latent regularization, possible target-content leakage,
and attention direction. All rows use the same data, architecture, optimizer,
mask sampler, pretraining budget, and seed. Cross-frame-only retains a
zero-weight same-frame forward for compute matching. Same-frame-only replaces
displaced targets with the source and removes nonzero $z(V)$ conditioning, so
it is a bundled endpoint rather than a loss-only ablation.

\begin{table*}[t]
\centering
\scriptsize
\setlength{\tabcolsep}{3.5pt}
\caption{Controlled ViT-B/16 ablations using a common pretraining seed (42).
Recognition columns report frozen top-1 accuracy (\%), averaged over three
probe seeds; ADE20K reports frozen-backbone mIoU. IN1K denotes ImageNet-1K. A
dash indicates that the transfer task was not evaluated.}
\label{tab:controlled-ablation}
\resizebox{\textwidth}{!}{%
\begin{tabular}{llccccc}
\toprule
Group & Configuration & IN1K & ADE20K & SSv2 joint-8 & UCF101 joint-8 & Diving48 joint-8 \\
\midrule
Reference & Final \ours{}
& $\mathbf{34.60{\pm}0.05}$ & $\mathbf{22.21}$ & $25.38{\pm}0.10$ & $56.60{\pm}0.38$ & $11.29{\pm}0.25$ \\
\midrule
Objective & Cross-frame only
& $32.26{\pm}0.05$ & $21.61$ & $\mathbf{29.77{\pm}0.18}$ & $\mathbf{58.15{\pm}0.21}$ & $11.54{\pm}0.16$ \\
& Same-frame only (bundled)
& $28.08{\pm}0.10$ & $21.89$ & $11.21{\pm}0.09$ & $47.83{\pm}0.21$ & $7.99{\pm}0.21$ \\
& No temporal offset ($\Delta{=}0$)
& $30.68{\pm}0.06$ & $20.50$ & $15.43{\pm}0.11$ & $52.85{\pm}0.26$ & $8.80{\pm}0.33$ \\
\midrule
Gradient path & Source stop-gradient (cross-frame)
& $29.75{\pm}0.12$ & $19.05$ & $13.99{\pm}0.12$ & $50.59{\pm}0.10$ & $10.41{\pm}0.36$ \\
\midrule
Condition & Zero $z(V)$
& $26.82{\pm}0.10$ & $20.92$ & $9.33{\pm}0.08$ & $44.94{\pm}0.12$ & $7.92{\pm}0.05$ \\
\midrule
Regularization & No $z(V)$ scale penalty
& $34.14{\pm}0.04$ & $21.79$ & $24.01{\pm}0.10$ & $54.11{\pm}0.18$ & $9.27{\pm}0.14$ \\
\midrule
Context access & Target content excluded from $z(V)$
& $33.01{\pm}0.04$ & $21.72$ & $23.65{\pm}0.11$ & $54.87{\pm}0.23$ & $10.90{\pm}0.18$ \\
& Random content excluded from $z(V)$
& $33.38{\pm}0.07$ & $21.69$ & $22.31{\pm}0.12$ & $52.08{\pm}0.18$ & $8.85{\pm}0.14$ \\
\midrule
Attention & Asymmetric attention
& $34.17{\pm}0.05$ & $22.06$ & $26.05{\pm}0.13$ & $55.21{\pm}0.30$ & $\mathbf{11.93{\pm}0.22}$ \\
\bottomrule
\end{tabular}%
}
\end{table*}

\paragraph{Objective and temporal conditioning.}
Cross-frame-only is strongest on joint action recognition, whereas adding the
same-frame task improves ImageNet and ADE20K, indicating a trade-off between
the two readouts. Stopping cross-frame gradients at the independently encoded
source features substantially reduces ImageNet, ADE20K, and joint action
transfer. The
gain from cross-frame prediction therefore depends on directly updating the
source-image representation, rather than training only the predictor and
video-context path. Replacing every temporal offset by zero or removing $z(V)$
lowers every reported task, showing that prediction uses both the requested
time and video-dependent context. This control removes direction and distance
together; it is not an isolated test of the sign alone.

\paragraph{Regularization, context completeness, and attention.}
The remaining rows of Table~\ref{tab:controlled-ablation} examine three design
choices. Removing the scale penalty lowers all five metrics by
$0.42$--$2.49$ points, supporting its use. For context access, target and
matched random exclusion both remove valid clip context.
This is consistent with computing the latents once, without target identities,
and sharing them across all predictions. Asymmetric attention has mixed effects
and is therefore not used. Appendix~\ref{app:additional-controls} provides the
detailed analysis and additional controls.

\begin{figure*}[t]
    \centering
    \includegraphics[width=0.88\textwidth]{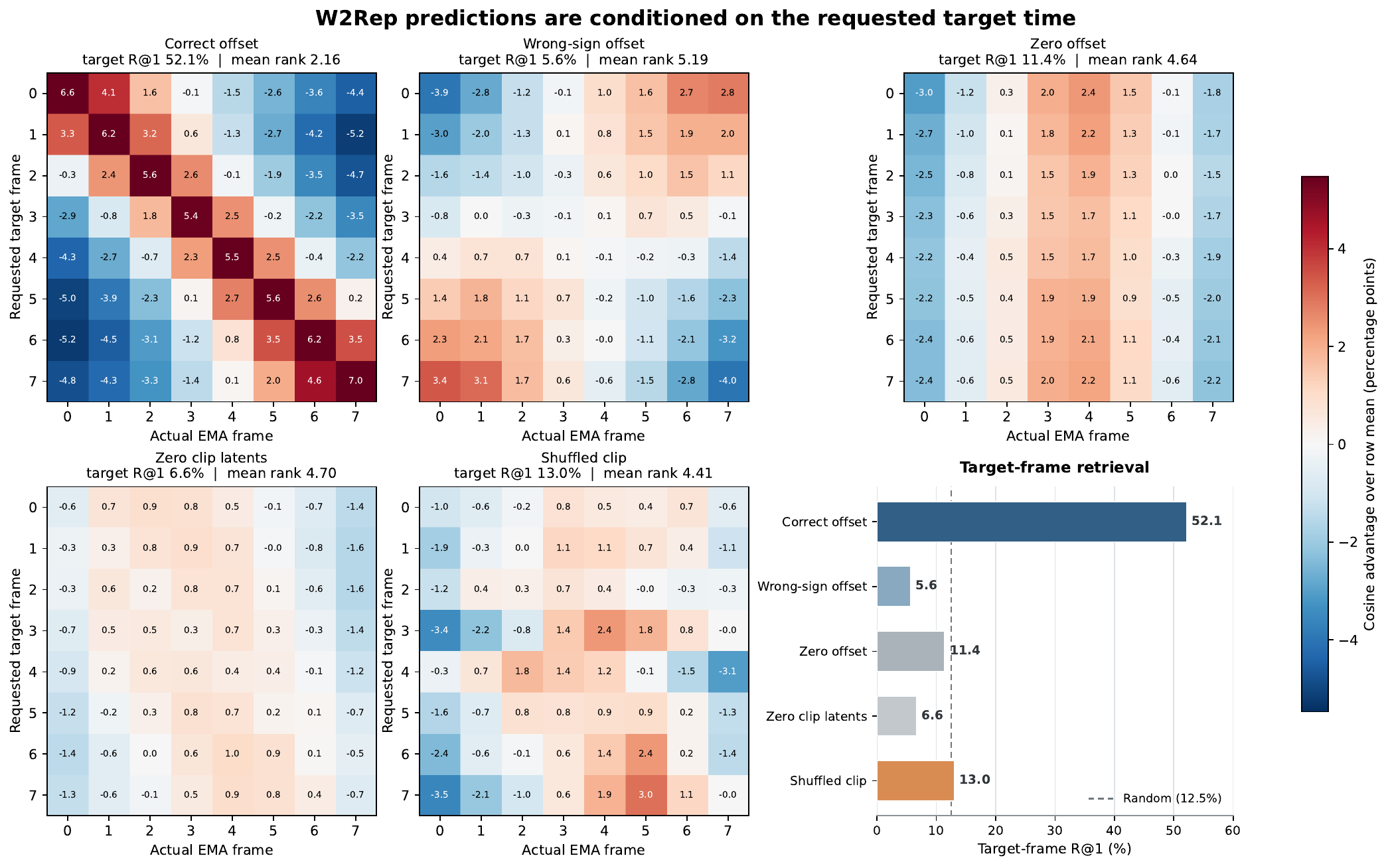}
    \caption{Temporal identification from predicted features on 512 SSv2
    validation videos. Each row requests one target time; each column compares
    the prediction with EMA features from one actual frame at the same masked
    query positions. Cells show cosine similarity relative to the mean of their
    row, in percentage points, and source-time requests are excluded from the
    retrieval statistics. Correct temporal conditioning produces a pronounced
    diagonal and retrieves the requested frame well above the $12.5\%$ random
    baseline. Perturbing the signed offset, clip order, or clip-dependent
    latents removes this structure.}
    \label{fig:temporal-prediction-matrix}
\end{figure*}

We further test temporal conditioning in
Fig.~\ref{fig:temporal-prediction-matrix}. With the correct offset, the
requested frame is retrieved first in $52.1\%$ of eight-way comparisons,
versus $12.5\%$ at random. Wrong-sign, zero-offset, zero-$z(V)$, and
shuffled-video interventions yield $5.6$--$13.0\%$, showing that predictions
depend jointly on the requested displacement and the ordered video evidence.

\paragraph{Content and use of the auxiliary clip latents.}
We inspect $z(V)$ from the final checkpoint to determine what context it makes
available to the predictor. A linear SSv2 head trained on pooled $z(V)$ reaches
$19.49\%$ on ordered clips, but $8.53\%$ after shuffling and $2.84\%$ when one
frame is repeated. Its nearest neighbor shares the action label in $11.30\%$
of ordered clips, compared with $0.86\%$ at random; static clips retain
$7.14\%$, showing that appearance and scene context also structure the latent
space. Replacing the matched $z(V)$ reduces prediction cosine from $0.790$ to
$0.453$ even when the donor has the same action label ($0.439$ for a different
label). Thus $z(V)$ combines temporal and visual context specific to the
current video. Appendix~\ref{app:z-content-analysis} provides the full
analysis. Additional protocols and diagnostic results appear in
Appendices~\ref{app:additional-controls}--\ref{app:latent-analysis}.

\section{Conclusion}
We introduced \ours{}, a masked cross-frame prediction objective that uses
temporal change to train a visual encoder. The source-image path must produce
features that support prediction across time, while the video path must gather
the complementary evidence supplied through $z(V)$. The retained ViT therefore
supports image-level and spatiotemporal video inference after all auxiliary
prediction components are removed. Across the current evaluations,
cross-frame, clip-conditioned pretraining improves semantic and action-related
transfer, while joint encoding adds further video utility; the advantage also
persists under matched end-to-end SSv2 fine-tuning. Retraining controls support
the utility of signed temporal offsets and do not indicate privileged access
to target content. Final-checkpoint diagnostics further show that the
auxiliary clip latents combine action- and order-sensitive information with
appearance and scene context, and that their useful contribution is strongly
specific to the observed video. Dense localization remains less competitive
than recognition under the current objective. Overall, watching how the world
changes can teach an encoder which visual states and interactions matter while
leaving a representation that remains useful for either images or videos.

\subsection*{AI use statement}
Generative AI tools were used to assist with language editing, code
development, experiment orchestration, and consistency checking. The authors
reviewed and verified the resulting text, code, analyses, and claims and take
full responsibility for the final content of this work.

\subsection*{Reproducibility statement}
The final objective and retained inference interface are specified in
Sections~\ref{sec:method} and~\ref{sec:shared-encoder}.
Section~\ref{sec:experiments} documents the comparison protocol and downstream
evaluations, while the appendices provide implementation details, additional
ablations, final-checkpoint temporal controls, and analyses of the auxiliary
clip latents. Exact
implementation artifacts and experiment configurations will accompany the
submission as supplementary material.

\bibliography{references}

@inproceedings{he2022masked,
  title={Masked autoencoders are scalable vision learners},
  author={He, Kaiming and Chen, Xinlei and Xie, Saining and Li, Yanghao and Doll{\'a}r, Piotr and Girshick, Ross},
  booktitle={2022 IEEE/CVF conference on computer vision and pattern recognition (CVPR)},
  pages={15979--15988},
  year={2022},
  organization={IEEE}
}

@inproceedings{assran2023self,
  title={Self-supervised learning from images with a joint-embedding predictive architecture},
  author={Assran, Mahmoud and Duval, Quentin and Misra, Ishan and Bojanowski, Piotr and Vincent, Pascal and Rabbat, Michael and LeCun, Yann and Ballas, Nicolas},
  booktitle={2023 IEEE/CVF Conference on Computer Vision and Pattern Recognition (CVPR)},
  pages={15619--15629},
  year={2023},
  organization={IEEE}
}

@inproceedings{caron2021emerging,
  title={Emerging properties in self-supervised vision transformers},
  author={Caron, Mathilde and Touvron, Hugo and Misra, Ishan and J{\'e}gou, Herv{\'e} and Mairal, Julien and Bojanowski, Piotr and Joulin, Armand},
  booktitle={2021 IEEE/CVF international conference on computer vision (ICCV)},
  pages={9630--9640},
  year={2021},
  organization={IEEE}
}

@article{caron2020unsupervised,
  title={Unsupervised learning of visual features by contrasting cluster assignments},
  author={Caron, Mathilde and Misra, Ishan and Mairal, Julien and Goyal, Priya and Bojanowski, Piotr and Joulin, Armand},
  journal={Advances in neural information processing systems},
  volume={33},
  pages={9912--9924},
  year={2020}
}

@article{tong2022videomae,
  title={Videomae: Masked autoencoders are data-efficient learners for self-supervised video pre-training},
  author={Tong, Zhan and Song, Yibing and Wang, Jue and Wang, Limin},
  journal={Advances in neural information processing systems},
  volume={35},
  pages={10078--10093},
  year={2022}
}

@inproceedings{pathak2017learning,
  title={Learning features by watching objects move},
  author={Pathak, Deepak and Girshick, Ross and Doll{\'a}r, Piotr and Darrell, Trevor and Hariharan, Bharath},
  booktitle={2017 IEEE Conference on Computer Vision and Pattern Recognition (CVPR)},
  pages={6024--6033},
  year={2017},
  organization={IEEE}
}

@article{bardes2024revisiting,
  title={Revisiting feature prediction for learning visual representations from video},
  author={Bardes, Adrien and Garrido, Quentin and Ponce, Jean and Chen, Xinlei and Rabbat, Michael and LeCun, Yann and Assran, Mahmoud and Ballas, Nicolas},
  journal={arXiv preprint arXiv:2404.08471},
  year={2024}
}

@inproceedings{bruce2024genie,
  title={Genie: Generative interactive environments},
  author={Bruce, Jake and Dennis, Michael D and Edwards, Ashley and Parker-Holder, Jack and Shi, Yuge and Hughes, Edward and Lai, Matthew and Mavalankar, Aditi and Steigerwald, Richie and Apps, Chris and others},
  booktitle={Forty-first international conference on machine learning},
  year={2024}
}

@article{bardes2023mc,
  title={Mc-jepa: A joint-embedding predictive architecture for self-supervised learning of motion and content features},
  author={Bardes, Adrien and Ponce, Jean and LeCun, Yann},
  journal={arXiv preprint arXiv:2307.12698},
  year={2023}
}

@inproceedings{mur2026v,
  title={V-jepa 2.1: Unlocking dense features in video self-supervised learning},
  author={Mur-Labadia, Lorenzo and Muckley, Matthew and Bar, Amir and Assran, Mido and Sinha, Koustuv and Rabbat, Mike and LeCun, Yann and Ballas, Nicolas},
  booktitle={European Conference on Computer Vision},
  pages={671--689},
  year={2026},
  organization={Springer}
}

@article{daithankar2026you,
  title={You Don't Need Strong Assumptions: Visual Representation Learning via Temporal Differences},
  author={Daithankar, Ninad and Gladstone, Alexi and LeCun, Yann and Ji, Heng},
  journal={arXiv preprint arXiv:2606.15956},
  year={2026}
}

@article{jang2024rsp,
  title={Visual representation learning with stochastic frame prediction},
  author={Jang, Huiwon and Kim, Dongyoung and Kim, Junsu and Shin, Jinwoo and Abbeel, Pieter and Seo, Younggyo},
  journal={arXiv preprint arXiv:2406.07398},
  year={2024}
}

@article{kim2026token,
  title={Token bottleneck: One token to remember dynamics},
  author={Kim, Taekyung and Han, Dongyoon and Heo, Byeongho and Park, Jeongeun and Yun, Sangdoo},
  journal={Advances in Neural Information Processing Systems},
  volume={38},
  pages={107455--107479},
  year={2026}
}

@inproceedings{chen2020simple,
  title={A simple framework for contrastive learning of visual representations},
  author={Chen, Ting and Kornblith, Simon and Norouzi, Mohammad and Hinton, Geoffrey},
  booktitle={International conference on machine learning},
  pages={1597--1607},
  year={2020},
  organization={PmLR}
}

@inproceedings{he2020momentum,
  title={Momentum contrast for unsupervised visual representation learning},
  author={He, Kaiming and Fan, Haoqi and Wu, Yuxin and Xie, Saining and Girshick, Ross},
  booktitle={2020 IEEE/CVF conference on computer vision and pattern recognition (CVPR)},
  pages={9726--9735},
  year={2020},
  organization={IEEE}
}

@article{grill2020bootstrap,
  title={Bootstrap your own latent-a new approach to self-supervised learning},
  author={Grill, Jean-Bastien and Strub, Florian and Altch{\'e}, Florent and Tallec, Corentin and Richemond, Pierre and Buchatskaya, Elena and Doersch, Carl and Avila Pires, Bernardo and Guo, Zhaohan and Gheshlaghi Azar, Mohammad and others},
  journal={Advances in neural information processing systems},
  volume={33},
  pages={21271--21284},
  year={2020}
}

@article{bao2021beit,
  title={Beit: Bert pre-training of image transformers},
  author={Bao, Hangbo and Dong, Li and Piao, Songhao and Wei, Furu},
  journal={arXiv preprint arXiv:2106.08254},
  year={2021}
}

@article{feichtenhofer2022masked,
  title={Masked autoencoders as spatiotemporal learners},
  author={Feichtenhofer, Christoph and Li, Yanghao and He, Kaiming and others},
  journal={Advances in neural information processing systems},
  volume={35},
  pages={35946--35958},
  year={2022}
}

@inproceedings{wei2022masked,
  title={Masked feature prediction for self-supervised visual pre-training},
  author={Wei, Chen and Fan, Haoqi and Xie, Saining and Wu, Chao-Yuan and Yuille, Alan and Feichtenhofer, Christoph},
  booktitle={2022 IEEE/CVF conference on computer vision and pattern recognition (CVPR)},
  pages={14648--14658},
  year={2022},
  organization={IEEE}
}

@inproceedings{misra2016shuffle,
  title={Shuffle and learn: unsupervised learning using temporal order verification},
  author={Misra, Ishan and Zitnick, C Lawrence and Hebert, Martial},
  booktitle={European conference on computer vision},
  pages={527--544},
  year={2016},
  organization={Springer}
}

@inproceedings{wang2019learning,
  title={Learning correspondence from the cycle-consistency of time},
  author={Wang, Xiaolong and Jabri, Allan and Efros, Alexei A},
  booktitle={2019 IEEE/CVF Conference on Computer Vision and Pattern Recognition (CVPR)},
  pages={2561--2571},
  year={2019},
  organization={IEEE}
}

@article{vaswani2017attention,
  title={Attention is all you need},
  author={Vaswani, Ashish and Shazeer, Noam and Parmar, Niki and Uszkoreit, Jakob and Jones, Llion and Gomez, Aidan N and Kaiser, {\L}ukasz and Polosukhin, Illia},
  journal={Advances in neural information processing systems},
  volume={30},
  year={2017}
}

@inproceedings{deng2009imagenet,
  title={Imagenet: A large-scale hierarchical image database},
  author={Deng, Jia and Dong, Wei and Socher, Richard and Li, Li-Jia and Li, Kai and Fei-Fei, Li},
  booktitle={2009 IEEE conference on computer vision and pattern recognition},
  pages={248--255},
  year={2009},
  organization={Ieee}
}

@article{russakovsky2015imagenet,
  title={Imagenet large scale visual recognition challenge},
  author={Russakovsky, Olga and Deng, Jia and Su, Hao and Krause, Jonathan and Satheesh, Sanjeev and Ma, Sean and Huang, Zhiheng and Karpathy, Andrej and Khosla, Aditya and Bernstein, Michael and others},
  journal={International journal of computer vision},
  volume={115},
  number={3},
  pages={211--252},
  year={2015},
  publisher={Springer}
}

@inproceedings{zhou2017scene,
  title={Scene parsing through ade20k dataset},
  author={Zhou, Bolei and Zhao, Hang and Puig, Xavier and Fidler, Sanja and Barriuso, Adela and Torralba, Antonio},
  booktitle={2017 IEEE conference on computer vision and pattern recognition (CVPR)},
  pages={5122--5130},
  year={2017},
  organization={IEEE}
}

@inproceedings{goyal2017something,
  title={The “something something” video database for learning and evaluating visual common sense},
  author={Goyal, Raghav and Kahou, Samira Ebrahimi and Michalski, Vincent and Materzynska, Joanna and Westphal, Susanne and Kim, Heuna and Haenel, Valentin and Fruend, Ingo and Yianilos, Peter and Mueller-Freitag, Moritz and others},
  booktitle={2017 IEEE international conference on computer vision (ICCV)},
  pages={5843--5851},
  year={2017},
  organization={IEEE}
}

@article{soomro2012ucf101,
  title={Ucf101: A dataset of 101 human actions classes from videos in the wild},
  author={Soomro, Khurram and Zamir, Amir Roshan and Shah, Mubarak},
  journal={arXiv preprint arXiv:1212.0402},
  year={2012}
}

@inproceedings{li2018resound,
  title={Resound: Towards action recognition without representation bias},
  author={Li, Yingwei and Li, Yi and Vasconcelos, Nuno},
  booktitle={European conference on computer vision},
  pages={520--535},
  year={2018},
  organization={Springer}
}

@inproceedings{xiao2018unified,
  title={Unified perceptual parsing for scene understanding},
  author={Xiao, Tete and Liu, Yingcheng and Zhou, Bolei and Jiang, Yuning and Sun, Jian},
  booktitle={European conference on computer vision},
  pages={432--448},
  year={2018},
  organization={Springer}
}

@article{loshchilov2017decoupled,
  title={Decoupled weight decay regularization},
  author={Loshchilov, Ilya and Hutter, Frank},
  journal={arXiv preprint arXiv:1711.05101},
  year={2017}
}

@article{bardes2021vicreg,
  title={Vicreg: Variance-invariance-covariance regularization for self-supervised learning},
  author={Bardes, Adrien and Ponce, Jean and LeCun, Yann},
  journal={arXiv preprint arXiv:2105.04906},
  year={2021}
}

@article{zhou2021ibot,
  title={ibot: Image bert pre-training with online tokenizer},
  author={Zhou, Jinghao and Wei, Chen and Wang, Huiyu and Shen, Wei and Xie, Cihang and Yuille, Alan and Kong, Tao},
  journal={arXiv preprint arXiv:2111.07832},
  year={2021}
}

@inproceedings{baevski2022data2vec,
  title={Data2vec: A general framework for self-supervised learning in speech, vision and language},
  author={Baevski, Alexei and Hsu, Wei-Ning and Xu, Qiantong and Babu, Arun and Gu, Jiatao and Auli, Michael},
  booktitle={International conference on machine learning},
  pages={1298--1312},
  year={2022},
  organization={PMLR}
}

@inproceedings{qian2021spatiotemporal,
  title={Spatiotemporal contrastive video representation learning},
  author={Qian, Rui and Meng, Tianjian and Gong, Boqing and Yang, Ming-Hsuan and Wang, Huisheng and Belongie, Serge and Cui, Yin},
  booktitle={2021 IEEE/CVF conference on computer vision and pattern recognition (CVPR)},
  pages={6960--6970},
  year={2021},
  organization={IEEE}
}

@inproceedings{wang2022long,
  title={Long-short temporal contrastive learning of video transformers},
  author={Wang, Jue and Bertasius, Gedas and Tran, Du and Torresani, Lorenzo},
  booktitle={2022 IEEE/CVF Conference on Computer Vision and Pattern Recognition (CVPR)},
  pages={13990--14000},
  year={2022},
  organization={IEEE}
}

@inproceedings{benaim2020speednet,
  title={Speednet: Learning the speediness in videos},
  author={Benaim, Sagie and Ephrat, Ariel and Lang, Oran and Mosseri, Inbar and Freeman, William T and Rubinstein, Michael and Irani, Michal and Dekel, Tali},
  booktitle={2020 IEEE/CVF Conference on Computer Vision and Pattern Recognition (CVPR)},
  pages={9919--9928},
  year={2020},
  organization={IEEE}
}

@inproceedings{wang2022bevt,
  title={Bevt: Bert pretraining of video transformers},
  author={Wang, Rui and Chen, Dongdong and Wu, Zuxuan and Chen, Yinpeng and Dai, Xiyang and Liu, Mengchen and Jiang, Yu-Gang and Zhou, Luowei and Yuan, Lu},
  booktitle={2022 IEEE/CVF Conference on Computer Vision and Pattern Recognition (CVPR)},
  pages={14713--14723},
  year={2022},
  organization={IEEE}
}

@inproceedings{girdhar2023omnimae,
  title={Omnimae: Single model masked pretraining on images and videos},
  author={Girdhar, Rohit and El-Nouby, Alaaeldin and Singh, Mannat and Alwala, Kalyan Vasudev and Joulin, Armand and Misra, Ishan},
  booktitle={2023 IEEE/CVF Conference on Computer Vision and Pattern Recognition (CVPR)},
  pages={10406--10417},
  year={2023},
  organization={IEEE}
}

@inproceedings{wang2023videomae,
  title={Videomae v2: Scaling video masked autoencoders with dual masking},
  author={Wang, Limin and Huang, Bingkun and Zhao, Zhiyu and Tong, Zhan and He, Yinan and Wang, Yi and Wang, Yali and Qiao, Yu},
  booktitle={2023 IEEE/CVF Conference on Computer Vision and Pattern Recognition (CVPR)},
  pages={14549--14560},
  year={2023},
  organization={IEEE}
}

@inproceedings{wang2023masked,
  title={Masked video distillation: Rethinking masked feature modeling for self-supervised video representation learning},
  author={Wang, Rui and Chen, Dongdong and Wu, Zuxuan and Chen, Yinpeng and Dai, Xiyang and Liu, Mengchen and Yuan, Lu and Jiang, Yu-Gang},
  booktitle={Proceedings of the IEEE/CVF conference on computer vision and pattern recognition},
  pages={6312--6322},
  year={2023}
}

@inproceedings{dwibedi2019temporal,
  title={Temporal cycle-consistency learning},
  author={Dwibedi, Debidatta and Aytar, Yusuf and Tompson, Jonathan and Sermanet, Pierre and Zisserman, Andrew},
  booktitle={2019 IEEE/CVF Conference on Computer Vision and Pattern Recognition (CVPR)},
  pages={1801--1810},
  year={2019},
  organization={IEEE}
}

@article{jabri2020space,
  title={Space-time correspondence as a contrastive random walk},
  author={Jabri, Allan and Owens, Andrew and Efros, Alexei},
  journal={Advances in neural information processing systems},
  volume={33},
  pages={19545--19560},
  year={2020}
}
\bibliographystyle{iclr2027_conference}

\appendix

\section{Discussion and Limitations}
\label{sec:discussion}
\paragraph{Why does one objective benefit both image and video inputs?}
The two encoder calls assign complementary roles to the cross-frame loss. The
independent source-image call must produce features that help predict another
moment. At the same time, the masked-video call must construct $z(V)$, which is
the predictor's only source of evidence from the other visible frames. A
useful $z(V)$ therefore requires the encoder to gather information across the
video rather than process its frames as unrelated images. Because both calls
use the same ViT, the loss trains one set of attention layers through both the
single-frame and multi-frame computations. The auxiliary tokens provide this
multi-frame training signal but are not themselves the downstream video
representation: after they are removed, the learned ViT can apply the same
attention layers to the patch tokens of one image or several frames. The joint
over independent gains in Table~\ref{tab:main-transfer} are consistent with
this mechanism, although they do not isolate the individual attention
interactions responsible for the gain.

\paragraph{What does prediction across time teach?}
Prediction across time relates different observations without requiring them
to have identical features. Treating frames as unrelated images discards their
connection, while directly enforcing temporal invariance can suppress changes
in pose, contact, and configuration. W2Rep instead makes a target feature
predictable from a source representation, visible video context, spatial
query, and temporal offset. Same-frame prediction anchors the representation
to spatial evidence, whereas cross-frame prediction asks it to remain useful
as the scene changes. We use \emph{temporally grounded visual state} to
describe this outcome: the representation is available from one image, but the
distinctions it preserves are shaped by observations across time. The local
similarities in Fig.~\ref{fig:cross-frame-patch-similarity} illustrate this
behavior but do not constitute an object-tracking result.

\paragraph{What does the auxiliary video context contribute?}
The diagnostics show that $z(V)$ combines temporal organization with
frame-visible context. Reordering a video reduces its action readout, whereas
static inputs retain substantial nearest-neighbor structure. Moreover, a
latent from the matched video is much more useful for prediction than a donor
latent, even when the donor has the same action label. These results indicate
that $z(V)$ supplies video-specific evidence about both how the observations
are organized and the particular objects, configuration, and scene in which
the change occurs.

\paragraph{Limitations and scope.}
The shared spatial coordinates used for cross-frame queries are a reference
system rather than an explicit correspondence mechanism, so object or camera
motion may displace relevant content. Dense transfer is also not a
demonstrated strength: W2Rep remains below VideoMAE on ADE20K at both scales.
Supervising only sampled final-layer masked positions is one possible reason,
but our experiments do not isolate it.

\section{Implementation and Evaluation Details}
\label{app:implementation-details}
This section specifies the training budgets, architectures, sampling rules,
and downstream protocols used for the results in the main paper.

\paragraph{Pretraining and comparison budgets.}
The resource controls separate video exposure from arithmetic cost because
neither quantity alone characterizes all objectives. All models are pretrained from
scratch on the same SSv2 split at $224$ resolution and use global batch 256.
For video-native methods, the primary protocol fixes a 100k-update training
horizon and eight-frame, stride-three clips. Thus \ours{}, VideoMAE, and V-JEPA
receive the same nominal number of clip samples, frames, and parameter updates.
For image- and frame-native methods, equal updates would process different
numbers of frames and training-time model calls. I-JEPA, TDV, and RSP are
therefore trained until their total profiled forward MACs match the \ours{}
budget at the corresponding backbone scale. Table~\ref{tab:resource-controls}
summarizes these rules.

Our MAC audit follows one convention for every method. It counts all modules
invoked by the optimized pretraining objective, including online encoders, EMA
target encoders, predictors, decoders, and auxiliary heads. It excludes the
backward pass, optimizer operations, and unsupported elementwise operators, so
it should be read as a reproducible forward-MAC accounting rather than a claim
of equal wall-clock time or complete training floating-point operations. At
ViT-B/16, the reference
budget is $3.22\times10^{18}$ forward MACs. External baselines retain their
method-specific masking, losses, and optimization schedules; changing these to
one common recipe would define a different method rather than only control its
resources.

\begin{table}[t]
\centering
\scriptsize
\setlength{\tabcolsep}{3.5pt}
\caption{Resource controls for the primary cross-method comparison. All
methods use the same SSv2 training split, input resolution, and global batch
size. The matching rule is fixed by the method's training interface, not by its
downstream result.}
\label{tab:resource-controls}
\begin{adjustbox}{max width=\linewidth}
\begin{tabular}{llll}
\toprule
Family & Methods & Training unit & Controlled resource \\
\midrule
Video-native
& \ours{}, VideoMAE, V-JEPA
& 8-frame clip
& 100k horizon; clips, frames, updates \\
Image/frame-native
& I-JEPA, TDV, RSP
& image or transition
& total profiled forward MACs \\
\bottomrule
\end{tabular}
\end{adjustbox}
\end{table}

No single protocol can simultaneously equalize both arithmetic cost and data
exposure when objectives have substantially different per-update costs. To
make this trade-off visible, Table~\ref{tab:videomae-compute-sensitivity}
additionally evaluates VideoMAE after matching the ViT-B/16 forward-MAC budget.
Because a VideoMAE update is cheaper, this checkpoint receives 799,367 updates
and consequently sees more video clips than either primary 100k-horizon row.
Additional compute improves VideoMAE, particularly for joint video encoding:
it exceeds \ours{} on joint SSv2 and UCF101 under this alternative protocol,
whereas \ours{} retains higher ImageNet and independent-frame SSv2 accuracy.
Accordingly, our main claim is representation utility under the declared
family-level controls, not compute-normalized dominance over every video
objective.

\begin{table*}[t]
\centering
\scriptsize
\setlength{\tabcolsep}{4pt}
\caption{Sensitivity to the resource-matching axis at ViT-B/16. All entries
report frozen top-1 accuracy (\%) and average three downstream-head seeds.
The primary protocol matches video exposure and the optimization horizon; the
additional VideoMAE row instead matches \ours{} at
$3.22\times10^{18}$ profiled forward MACs.}
\label{tab:videomae-compute-sensitivity}
\begin{adjustbox}{max width=\textwidth}
\begin{tabular}{llccccccc}
\toprule
Method & Resource control & ImageNet & \multicolumn{2}{c}{SSv2} &
\multicolumn{2}{c}{UCF101} & \multicolumn{2}{c}{Diving48} \\
\cmidrule(lr){4-5}\cmidrule(lr){6-7}\cmidrule(lr){8-9}
& & & Ind.-8 & Joint-8 & Ind.-8 & Joint-8 & Ind.-8 & Joint-8 \\
\midrule
VideoMAE & 100k video horizon
& $26.70{\pm}0.10$ & $8.98{\pm}0.04$ & $24.27{\pm}0.07$
& $46.29{\pm}0.15$ & $53.87{\pm}0.19$ & $8.00{\pm}0.11$ & $10.24{\pm}0.33$ \\
VideoMAE & Forward-MAC matched
& $31.01{\pm}0.10$ & $11.33{\pm}0.08$ & $30.46{\pm}0.03$
& $52.08{\pm}0.39$ & $59.56{\pm}0.49$ & $9.37{\pm}0.03$ & $11.56{\pm}0.21$ \\
\ours{} & 100k video horizon
& $34.60{\pm}0.05$ & $12.64{\pm}0.03$ & $25.38{\pm}0.10$
& $51.40{\pm}0.28$ & $56.60{\pm}0.38$ & $10.08{\pm}0.26$ & $11.29{\pm}0.25$ \\
\bottomrule
\end{tabular}
\end{adjustbox}
\end{table*}

\paragraph{\ours{} architecture and optimization.}
The two model scales differ in encoder and predictor depth while sharing the
same auxiliary-context design and optimization recipe. ViT-B/16 uses a
12-block, width-768 encoder with 12 heads and a six-block,
width-384 predictor with 12 heads. ViT-L/16 uses a 24-block, width-1024 encoder
with 16 heads and a 12-block, width-384 predictor with 12 heads. Both use
$K{=}16$ auxiliary clip latents. We train with AdamW
\citep{loshchilov2017decoupled}, global batch size 256,
$(\beta_1,\beta_2)=(0.9,0.999)$, and bfloat16 arithmetic. The learning rate
warms from $2\times10^{-4}$ to $10^{-3}$ over 10k updates and then follows cosine decay to
$10^{-6}$. Weight decay is scheduled from $0.04$ to $0.4$, gradients are
clipped at norm $1.0$, and EMA momentum increases linearly from $0.996$ to
$1.0$.

\paragraph{\ours{} masking and sampling.}
W2Rep uses one spatial mask across the video and samples target frames on both
sides of a uniformly selected source frame. We sample four target blocks with
area scale $[0.15,0.20]$ and aspect-ratio
range $[0.75,1.5]$. A large encoder block with area scale $[0.85,1.0]$ is
sampled from the complement of all target blocks, so the visible locations do
not overlap the prediction queries. The same spatial mask and geometric
augmentation are applied to every frame. Each training sample contains eight
RGB frames at stride three. We draw the source index uniformly and sample three
target frames without replacement from the remaining frames, allowing both
positive and negative time offsets. Accordingly, $\Delta$ is the index
difference in this sampled eight-frame sequence, and one unit corresponds to
three frames in the original video.

External baselines retain their method-specific masking, losses, EMA,
optimization, and frame/tubelet tokenization except for disclosed input
adaptations. VideoMAE uses eight frames at stride three, tubelet size
two, 90\% tube masking, and a four-block decoder; its audit includes the
encoder, encoder-to-decoder projection, decoder, and pixel head. V-JEPA retains
its native latent-prediction objective. RSP-B/16 retains its stochastic
future-representation and auxiliary masked-reconstruction objectives, uses its
75\% reconstruction mask, and is matched to \ours{}-B/16 by profiled forward
MACs; its retained image encoder is evaluated with independent-frame readouts.
Under the common frozen-backbone ADE20K protocol, its final 160k-iteration
decoder obtains $17.84$ mIoU, $66.06$ overall pixel accuracy (aAcc), and
$24.11$ mean class accuracy (mAcc).
For TDV, we report the frozen EMA frame
encoder from the completed two-frame run. Each sample contains one transition,
and gradient accumulation gives a global transition batch of 256.

\paragraph{Frozen downstream protocols.}
Frozen evaluation retains only the encoder and distinguishes separate image
calls from joint video encoding. For image-native encoders and \ours{}, one
image is processed in one $T{=}1$ call. VideoMAE and V-JEPA use temporal
tubelets of size two, so their image readout repeats the same RGB image once to
form a single tubelet. For independent-$8$, this adaptation is applied in eight
separate Transformer calls and the resulting descriptors are averaged; frames
from different times never interact inside the encoder. For joint-$8$, the
eight actual RGB frames are processed together as four tubelets. The native
positional encoding of each video encoder is evaluated on the resulting token
sequence; no pretrained projection or Transformer weight is modified.
Independent-$8$ uniformly samples eight stored RGB frames, encodes them
separately, and averages spatially pooled descriptors. Joint-$8$ sends exactly
the same frames through one encoder call before global pooling. The auxiliary
tokens are disabled for every standard downstream readout. ViT-B heads use
768-dimensional descriptors and ViT-L heads use 1024-dimensional descriptors.

Each downstream dataset uses a fixed protocol without validation-based
checkpoint selection. ImageNet uses all $1.28$M training images and patch-mean
frozen features.
ADE20K uses a frozen-backbone UPerNet trained for 160k iterations at
$512{\times}512$; I-JEPA-L, VideoMAE-L, V-JEPA-L, and \ours{}-L use
this same protocol. SSv2 uses the full 174-class split. UCF101 uses official
split~1, a deterministic eight-frame cache, one spatial view, and standardized
linear heads trained for 50 epochs. Diving48 uses the cleaned v2 split with 47
active classes and the same frame cache and head schedule. Recognition values
with $\pm$ average fixed final-epoch heads with seeds $42,43,44$ rather than
selecting a test-set peak. Temporal controls freeze heads trained on ordered
descriptors before perturbing validation input; uncertainty uses paired
video- or clip-level bootstrap resampling.

\paragraph{SSv2 full fine-tuning.}
The full-fine-tuning comparison uses one matched downstream recipe for all
ViT-B encoders. Each model is initialized from its designated primary
pretraining checkpoint and jointly updates the encoder and a linear 174-way classifier for
50 epochs. All methods use AdamW, global batch 256, base
learning rate $5\times10^{-4}$, five warm-up epochs, cosine decay to
$10^{-6}$, weight decay $0.05$, layer-wise learning-rate decay $0.75$, and
label smoothing $0.1$. Training samples one frame from each of eight temporal
segments and applies one spatial crop consistently across the clip; validation
uses eight uniformly spaced frames and a direct $224{\times}224$ resize. We
report the fixed epoch-50 checkpoint with no validation-based selection.
I-JEPA averages eight independently encoded frames; V-JEPA, VideoMAE, and
\ours{} jointly encode them. Predictor and auxiliary tokens are absent from
the \ours{} downstream graph.

\begin{table}[t]
\centering
\scriptsize
\setlength{\tabcolsep}{3.5pt}
\caption{SSv2 full fine-tuning with ViT-B/16. All methods use the same data,
augmentations, optimization schedule, and fixed epoch-50 reporting rule.
Results are validation accuracy (\%) from each method's designated primary
checkpoint. I-JEPA encodes frames separately; all other methods encode them
together.}
\label{tab:ssv2-fullft}
\begin{tabular}{lcccc}
\toprule
Method & Encoding & Top-1 & Top-5 & Macro Acc. \\
\midrule
I-JEPA   & Ind.-8   & $12.23$ & $28.43$ & $8.19$ \\
V-JEPA   & Joint-8  & $45.48$ & $70.44$ & $37.25$ \\
VideoMAE & Joint-8  & $55.16$ & $79.13$ & $48.30$ \\
\ours{}  & Joint-8  & $\mathbf{58.77}$ & $\mathbf{81.84}$ & $\mathbf{51.79}$ \\
\bottomrule
\end{tabular}
\end{table}

\section{Additional Ablation Results}
\label{app:additional-controls}
This section expands the main ablation study with dense-transfer metrics,
pretraining-seed variation, and the auxiliary-latent capacity sweep.

\subsection{Dense-transfer details}
Dense-transfer results test whether the main ablation trends extend beyond
recognition. Table~\ref{tab:ade-objective-controls} expands the ADE20K column of the main
ablation table with pixel and class accuracy. The target-exclusion control
prevents $z(V)$ from reading visible patches from the sampled target frames;
the matched random control removes the same number of non-target frames.

\begin{table}[t]
\centering
\scriptsize
\setlength{\tabcolsep}{4pt}
\caption{ADE20K frozen-backbone transfer for objective and auxiliary-latent
controls. All controlled pretraining runs use the common seed 42, and all
decoders use their final 160k-iteration checkpoint. $\Delta$ mIoU is measured
relative to final \ours{}; aAcc and mAcc denote overall pixel accuracy and mean
class accuracy, respectively.}
\label{tab:ade-objective-controls}
\begin{adjustbox}{max width=\linewidth}
\begin{tabular}{lcccc}
\toprule
Configuration & mIoU & $\Delta$ mIoU & aAcc & mAcc \\
\midrule
Final \ours{}                  & $\mathbf{22.21}$ & --      & $\mathbf{69.44}$ & $\mathbf{29.61}$ \\
Cross-frame only               & $21.61$ & $-0.60$ & $69.02$ & $28.71$ \\
No temporal offset ($\Delta{=}0$) & $20.50$ & $-1.71$ & $68.25$ & $27.19$ \\
Source stop-gradient (cross-frame) & $19.05$ & $-3.16$ & $67.09$ & $25.54$ \\
No $z(V)$ scale penalty        & $21.79$ & $-0.42$ & $69.63$ & $29.10$ \\
Target content excluded from $z(V)$ & $21.72$ & $-0.49$ & $69.17$ & $28.99$ \\
Random content excluded from $z(V)$ & $21.69$ & $-0.52$ & $69.11$ & $28.86$ \\
\bottomrule
\end{tabular}
\end{adjustbox}
\end{table}

Stopping the cross-frame gradient at the source representation has the largest
dense-transfer cost, reinforcing that this loss must directly shape the image
features rather than only train the predictor and video-context path. Removing
the temporal-offset condition causes the next-largest drop. The scale penalty
also improves each task's primary metric, including $0.42$ mIoU on ADE20K. It
controls the magnitude of the predictor's video condition, although these
results do not distinguish an optimization effect from reduced reliance on
that branch.

The context-exclusion controls require a different interpretation. The same
auxiliary latents are reused for every target and are computed without target
identities or offsets, which encourages a reusable clip summary. Nevertheless,
visible target-frame regions are legitimate context for masked prediction.
Excluding them therefore makes the conditioning clip incomplete in addition
to removing a possible target-specific route.

\subsection{Auxiliary-latent capacity}
The sweep in Table~\ref{tab:k-capacity} favors $K{=}16$ as a balanced
empirical choice: it gives the
strongest ImageNet, SSv2, and UCF101 readouts. $K{=}4$ remains close on SSv2
but loses 2.02 points on ImageNet and 2.47 on UCF101 joint-$8$; $K{=}64$ does
not improve these readouts and loses 4.14 on SSv2 joint-$8$. The non-monotonic
sweep does not establish an information bottleneck.

\begin{table*}[t]
\centering
\scriptsize
\setlength{\tabcolsep}{3.5pt}
\caption{Auxiliary-latent capacity sweep using the common controlled seed 42
and 100k updates. Values
are top-1 accuracy (\%), averaged over three downstream heads.}
\label{tab:k-capacity}
\begin{adjustbox}{max width=\linewidth}
\begin{tabular}{cccccccc}
\toprule
$K$ & IN1K & SSv2 Ind.-8 & SSv2 Joint-8 & UCF101 Ind.-8 & UCF101 Joint-8 & Diving48 Ind.-8 & Diving48 Joint-8 \\
\midrule
$4$  & $32.58{\pm}0.12$ & $12.20{\pm}0.05$ & $25.13{\pm}0.07$ & $48.41{\pm}0.12$ & $54.14{\pm}0.12$ & $10.10{\pm}0.36$ & $\mathbf{11.51{\pm}0.41}$ \\
$16$ & $\mathbf{34.60{\pm}0.05}$ & $\mathbf{12.64{\pm}0.03}$ & $\mathbf{25.38{\pm}0.10}$ & $\mathbf{51.40{\pm}0.28}$ & $\mathbf{56.60{\pm}0.38}$ & $10.08{\pm}0.26$ & $11.29{\pm}0.25$ \\
$64$ & $34.24{\pm}0.03$ & $12.25{\pm}0.06$ & $21.25{\pm}0.11$ & $50.16{\pm}0.22$ & $54.94{\pm}0.29$ & $\mathbf{10.47{\pm}0.08}$ & $11.12{\pm}0.35$ \\
\bottomrule
\end{tabular}
\end{adjustbox}
\end{table*}

\section{Temporal Controls}
\label{app:temporal-controls}
This section tests whether the retained encoder and frozen predictor respond to
the temporal order, requested displacement, and video-specific conditioning.

\subsection{Final-checkpoint representation sensitivity}
The retained joint encoder is evaluated under input-order interventions while
its linear heads remain fixed. Table~\ref{tab:final-order-controls} uses the
designated 100k-update W2Rep checkpoint. Three epoch-50 linear heads trained on ordered
SSv2 joint-$8$ descriptors are frozen and reused for every condition. The same
eight decoded RGB frames from each of the $24{,}777$ validation videos are
present in the ordered, reversed, and fixed-shuffled conditions; only their
order changes. The static-repeat control repeats sampled frame 4 eight times
and therefore removes frame diversity in addition to temporal order.

\begin{table}[t]
\centering
\scriptsize
\setlength{\tabcolsep}{4pt}
\caption{Final-checkpoint SSv2 order sensitivity. Top-1 values are mean and
standard deviation over the same three frozen heads. Drops are paired against
ordered input at the video level.}
\label{tab:final-order-controls}
\begin{tabular}{lcc}
\toprule
Encoder input & Top-1 (\%) & Ordered drop \\
\midrule
Ordered        & $25.38{\pm}0.10$ & -- \\
Reversed       & $13.51{\pm}0.04$ & $11.88$ \\
Fixed shuffled & $8.98{\pm}0.06$  & $16.40$ \\
Static repeat  & $4.24{\pm}0.09$  & $21.14$ \\
\bottomrule
\end{tabular}
\end{table}

The fixed-head results show substantial sensitivity to both temporal order and
frame diversity. Across 5,000 paired video-level bootstrap replicates, the $95\%$ confidence
intervals for ordered-minus-reversed, ordered-minus-shuffled, and
ordered-minus-static are $[11.43,12.32]$, $[15.91,16.89]$, and
$[20.61,21.66]$ points. This establishes order sensitivity of the retained
joint encoder and its fixed heads; it is not a retraining ablation and does not
measure how much temporal order caused the learned representation.

\subsection{Final-checkpoint predictor sensitivity}
Frozen-predictor interventions isolate the conditioning variables used for
cross-frame prediction. Table~\ref{tab:final-objective-controls} evaluates the
complete
SSv2 validation set. Each video uses the same source frame, three displaced
targets, spatial masks, EMA target features, and geometric transform in every
condition. The correct condition uses the ordered clip, signed temporal offset,
and clip-dependent $z(V)$. Each control changes only the listed predictor
condition; in particular, zero offset retains the nonzero $z(V)$ rather than
invoking the training-time same-frame rule.

\begin{table}[t]
\centering
\scriptsize
\setlength{\tabcolsep}{3.5pt}
\caption{Frozen-predictor interventions on the final checkpoint. Relative
increase is measured against the correct-condition Smooth L1 loss.}
\label{tab:final-objective-controls}
\begin{adjustbox}{max width=\linewidth}
\begin{tabular}{lccc}
\toprule
Predictor condition & Smooth L1 $\downarrow$ & Relative increase & Cosine $\uparrow$ \\
\midrule
Correct ordered clip + offset & $0.131518$ & --        & $0.786827$ \\
Fixed-shuffled clip           & $0.148446$ & $12.87\%$ & $0.740647$ \\
Wrong-sign offset             & $0.159054$ & $20.94\%$ & $0.710809$ \\
Zero offset, keep $z(V)$      & $0.147933$ & $12.48\%$ & $0.742612$ \\
Zero $z(V)$, keep offset      & $0.168696$ & $28.27\%$ & $0.685907$ \\
\bottomrule
\end{tabular}
\end{adjustbox}
\end{table}

Every intervention reliably increases prediction loss relative to the correct
condition. Paired bootstrap intervals for the absolute loss increase over the correct
condition are $[0.016711,0.017154]$ for shuffled clips,
$[0.027204,0.027893]$ for wrong-sign offsets,
$[0.016193,0.016637]$ for zero offsets, and
$[0.036917,0.037436]$ for zero $z(V)$. The trained predictor therefore uses
clip order, signed offsets, and clip-dependent conditioning. These are
inference-time interventions on one trained model and do not substitute for
pretraining separate models under wrong-order or wrong-offset objectives.

\section{Analysis of the Auxiliary Clip Latents}
\label{app:latent-analysis}
This section characterizes the information available through $z(V)$ and tests
how that information affects prediction within and beyond the pretraining
domain.

\subsection{Final-checkpoint content and swap diagnostics}
\label{app:z-content-analysis}
The final-checkpoint diagnostics separate decodable content from the functional
effect of matched video context. We analyze the auxiliary clip latents of the
designated 100k-update checkpoint on the complete SSv2 train and validation
splits. Because the
online encoder supplied $z(V)$ to the predictor during pretraining, we inspect
that branch rather than the EMA target used for standard downstream probes.
Each descriptor averages the $K{=}16$ normalized latent tokens produced from a
masked eight-frame clip. Clip sampling, spatial masking, and center cropping
are fixed deterministically for this analysis.

\begin{figure*}[t]
    \centering
    \includegraphics[width=0.88\textwidth]{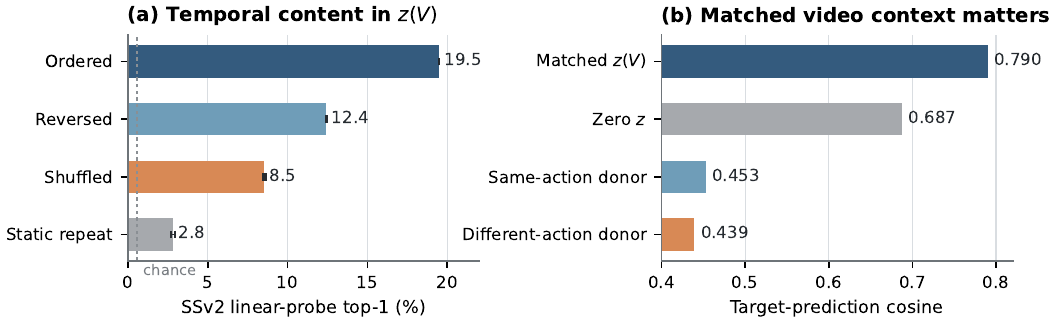}
    \caption{Content and functional diagnostics of $z(V)$ at the final
    checkpoint. (a) An SSv2 linear head trained on ordered clip latents is
    evaluated without refitting; accuracy falls when the same frames are
    reversed or shuffled, and further when one frame is repeated. Error bars
    show standard deviation over three head seeds. (b) We hold the source,
    targets, masks, and offsets fixed and change only $z(V)$. Prediction is
    strongest with context from the matched video; a donor from another video
    remains much less useful even when it has the same action label.}
    \label{fig:z-content-summary}
\end{figure*}

\paragraph{Decodable content and temporal controls.}
An ordered linear readout is sensitive to both the set of observed frames and
their temporal organization. We train a linear 174-way SSv2 head on ordered training descriptors for a fixed
50 epochs and reuse it unchanged for all validation conditions. Ordered,
reversed, and fixed-shuffled inputs contain exactly the same RGB frames;
repeated-static input uses sampled frame 4 at all eight positions. As shown in
Fig.~\ref{fig:z-content-summary}(a), ordered $z(V)$ reaches
$19.49{\pm}0.02\%$ top-1 and $43.53{\pm}0.12\%$ top-5 over three head seeds.
The same head reaches $12.43\%$, $8.53\%$, and $2.84\%$ after reversal,
shuffling, and static repetition. Paired ordered-minus-control top-1 intervals
are $[6.67,7.45]$, $[10.53,11.40]$, and $[16.18,17.12]$ points,
respectively. Thus the action-related information accessible to an ordered
linear readout depends on both multiple observations and their temporal
organization.

Nearest-neighbor geometry remains substantially organized by appearance and
scene context. The nearest ordered descriptor shares its SSv2 action label in
$11.30\%$ of validation videos,
compared with $0.86\%$ for an empirical random video. Reversal and shuffling
give $11.13\%$ and $10.24\%$, while repeated-static input remains at
$7.14\%$. The persistence of structure under static input shows that
frame-visible appearance and scene context remain substantial organizing cues.
Meanwhile, the pooled descriptor changes with the input: its mean cosine with
ordered $z(V)$ is $0.856$ for reversed, $0.858$ for shuffled, and $0.670$ for
static clips. The fixed-head accuracy and nearest-neighbor statistics measure
different properties: the former depends on alignment with the learned
ordered decision boundary, whereas the latter measures neighborhood structure
within each controlled representation space.

\paragraph{Matched-context intervention.}
Matched video context is substantially more useful to the predictor than a
donor selected only by action label. We hold the primary video's source frame, three target frames, spatial
masks, EMA targets, and signed offsets fixed and replace only the predictor's
$z(V)$. The donor is either the next distinct validation video with the same
SSv2 action label or a deterministically selected video with a different
label. The correct latent gives target-feature cosine $0.790$ and Smooth L1
loss $0.130$. A same-label donor gives $0.453/0.244$, a different-label donor
gives $0.439/0.248$, and zero $z$ gives $0.687/0.168$. The same-label donor is
$0.0134$ cosine better than the different-label donor (paired $95\%$ interval
$[0.0121,0.0146]$), revealing a small action-category component. The much
larger gap to the correct latent shows that the predictor primarily needs
context matched to the particular video---including its current appearance,
configuration, and scene---rather than an interchangeable class-level code.
An inconsistent donor being worse than zero reflects conflicting conditioning,
not intrinsically harmful information in the donor video.

Taken together, these measurements characterize $z(V)$ as video-specific
predictive context that combines action and interaction cues, temporal
organization, and the appearance and scene in which the interaction occurs.
The static control removes both temporal evolution and multi-frame diversity,
whereas shuffling is the cleaner order-only intervention. Because the action
probe is evaluated on the pretraining dataset, it is a content diagnostic
rather than an additional transfer result.

\subsection{Qualitative cross-domain analysis on EgoDex}
\label{sec:egodex-qualitative}
The EgoDex visualization examines whether action direction or recording
context dominates frozen nearest-neighbor similarity outside the pretraining
domain. This diagnostic uses frozen W2Rep representations but is not an
official EgoDex benchmark or an additional transfer claim.
For a query episode, we retrieve a different episode from the same coarse
action family using either the retained encoder representation or the
auxiliary video context $z(V)$. Exact-task matches are excluded, and each retrieved episode is
displayed as four chronological frames so that action direction can be judged
from the sequence rather than inferred from one image.

\begin{figure*}[t]
    \centering
    \includegraphics[width=0.98\textwidth]{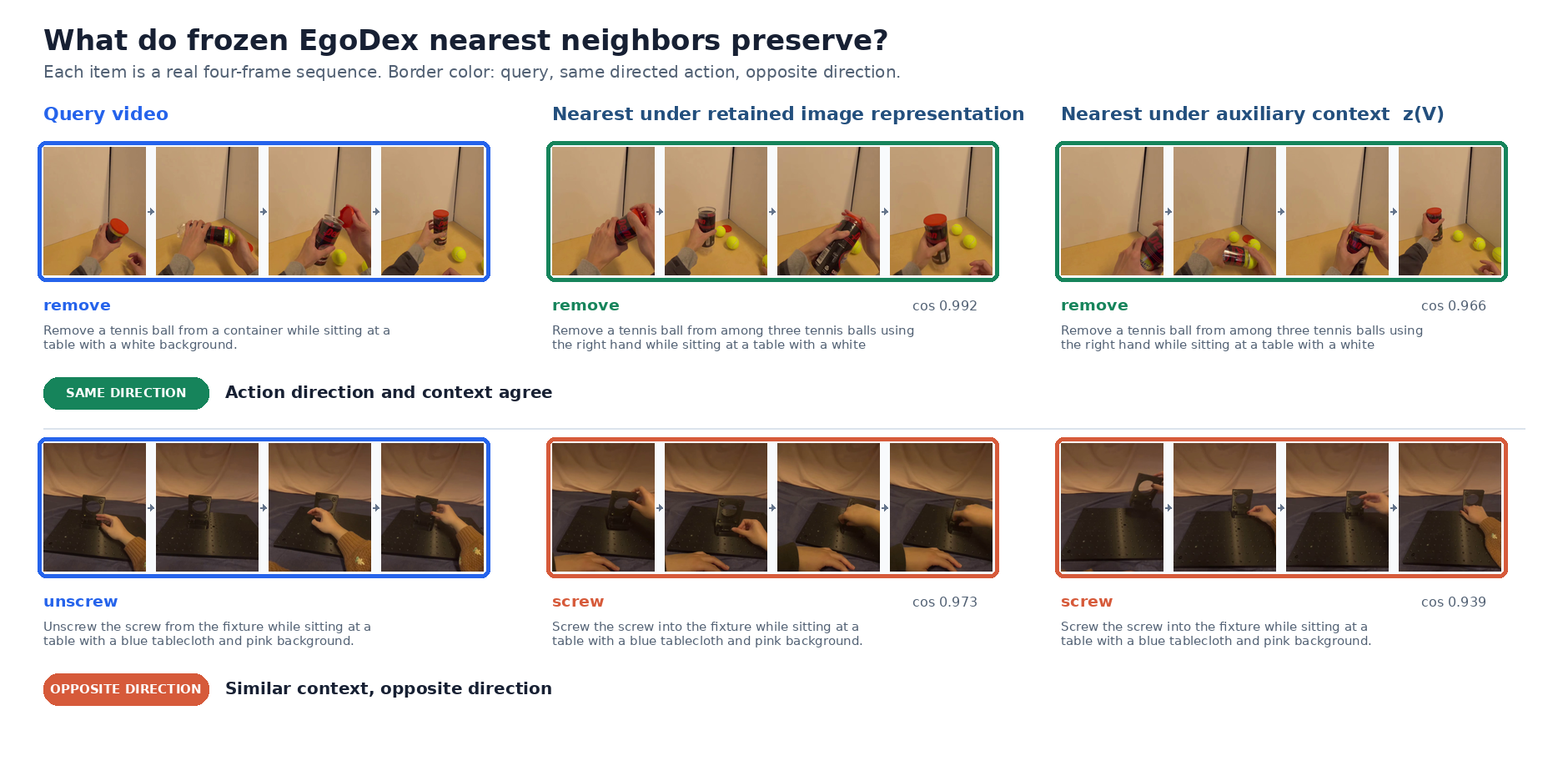}
    \caption{Qualitative nearest-neighbor retrieval on EgoDex with frozen
    \ours{} representations. Each item shows
    four chronological frames from a real episode. Green borders denote the
    same directed action as the query, while orange borders denote its inverse.
    The upper example preserves both action direction and visual context. In
    the lower example, the near-identical object and recording setup outweigh
    action direction for both the retained encoder representation and the
    auxiliary video context $z(V)$.}
    \label{fig:egodex-qualitative}
\end{figure*}

EgoDex neighbors exhibit both action-compatible and context-dominated
similarity. Figure~\ref{fig:egodex-qualitative} illustrates these behaviors. In
the first example, the nearest neighbors match the query's action direction as
well as its object and recording context. In the second, both representations
retrieve the inverse action because the fixture, viewpoint, and background are
nearly identical. The examples suggest that the retained encoder representation and $z(V)$ organize videos by
a mixture of interaction and visual context; action direction can be retained,
but it need not dominate similarity when the recording setup is highly
matched. This is consistent with the role of $z(V)$ as video-specific context
for prediction rather than a standalone downstream representation.

\end{document}